\documentclass{llncs}

\usepackage{makeidx}  
\usepackage{url,color,algorithmic,algorithm}
\usepackage{graphicx}

\usepackage{subfigure}
\usepackage{amsmath,mathtools}

\begin{document}

\frontmatter          

\pagestyle{headings}  
\addtocmark{Skill model} 

\title{Computational Red Teaming in a Sudoku Solving Context: Neural Network Based Skill Representation and Acquisition}

\titlerunning{Sudoku: skill representation and acquisition}  

\author{George Leu \and Hussein Abbass}

\authorrunning{Leu, and Abbass} 

\tocauthor{George Leu, and Hussein Abbass}

\institute{School of Engineering and Information Technology\\
University of New South Wales (Canberra Campus), Australia\\
\email{(G.Leu,H.Abbass)@adfa.edu.au}}

\maketitle              

\begin{abstract}
In this paper we provide an insight into the skill representation, where skill representation is seen as an essential part of the skill assessment stage in the Computational Red Teaming process. Skill representation is demonstrated in the context of Sudoku puzzle, for which the real human skills used in Sudoku solving, along with their acquisition, are represented computationally in a cognitively plausible manner, by using feed-forward neural networks with back-propagation, and supervised learning. The neural network based skills are then coupled with a hard-coded constraint propagation computational Sudoku solver, in which the solving sequence is kept hard-coded, and the skills are represented through neural networks. The paper demonstrates that the modified solver can achieve different levels of proficiency, depending on the amount of skills acquired through the neural networks. Results are encouraging for developing more complex skill and skill acquisition models usable in general frameworks related to the skill assessment aspect of Computational Red Teaming.

\keywords{neural network, domain propagation, skill acquisition, supervised learning}
\end{abstract}

\section{Introduction}

In Computational Red Teaming (CRT) a Red agent takes actions to challenge a Blue agent, with a variety of purposes. In the cognitive domain, one of these purposes, which generated an intense interest in the scientific community in recent years, is to force a human Blue agent to improve its skills. This process involves two major aspects. First, the Red must find the proper ways of action for challenging the Blue; this is the \textit{task probing}. Second, in order to find those ways of action, the Red must first assess Blue's skills for finding its weaknesses and hence, potential directions of improvement. This second aspect is the \textit{skill assessment} aspect, in which the representation of Blue's skills is essential.

In this paper we apply the CRT to Sudoku puzzle and we focus on the representation of the skills used for solving a Sudoku game. We investigate the Sudoku literature in order to establish what are the skills that humans apply to solve the puzzles, and then we create their computational representation, in a manner that is cognitively plausible. We use feed-forward neural networks (NN) to represent the skills, and we model the skill acquisition process through supervised learning and back propagation. The NN-based skills are then embedded into a classic hard-coded constraint propagation Sudoku solver, endowing it with the ability to learn Sudoku skills through training. While the Sudoku solving sequence remains hard-coded, the computational solver uses at each predefined step the pattern recognition capability of the neural networks, and thus, its proficiency varies based on the skills embedded in its structure. In order to demonstrate this we use two skill setups: a first one in which the neural networks can only detect the existence of a favourable pattern on te Sudoku board, and a second one in which the pattern can be both detected and localised. Simulation results show how the realistic skill-based solver can achieve different levels of proficiency in solving Sudoku in the two setups, with a higher level of proficiency reached for the first skill setup.

The paper is organised as follows. The second section presents the existing computational approaches on solving Sudoku and draws a conclusion on the lack of skill-based computational solvers. The third section shows how we choose from the range of human skills used in Sukodu solving, in order to transfer them into the proposed skill representation and acquisition model. The fourth section describes the methodology used for modelling the skills and the NN-based skill acquisition process. The fifth section presents and discusses the results of the experiments. Last section concludes the study and summarises the main findings.

\section{Background on computational Sudoku solving}

The existing computational Sudoku solvers focus mostly on reducing the implementation and computational complexity, and on solving the puzzle as a search/optimisation problem, without Sudoku domain-specific knowledge or concerns about the cognitive plausibility.

From a computational perspective several Sudoku solvers have been reported in the literature. The simplest, but also the least effective is the backtracking solver, a brute force method that uses the full space of possible grids and performs a backtracking-based depth-first search through the resultant solution tree \cite{Berggren2012}. Another simple solver is  the ``pencil and paper" algorithm \cite{Crook2009} which visits cells in the grid and generates on the fly a search tree.

In a strict mathematical view, the general $n\times n$ formulation of Sudoku is considered a non-deterministic polynomial time (NP)  problem. An open question still exists in the literature on whether Sudoku belongs or not to the subclass of NP-complete problems, however more authors seem to be on the NP-completeness side \cite{Yato2003,Yato2003a,Weber2005,ErcseyRavasz2012,Goldberg2015}. Yato \cite{Yato2003} and Yato and Seta \cite{Yato2003a} first demonstrated that the generalised $n\times n$ Sudoku problem can be solved in polynomial time. Later, another approach \cite{ErcseyRavasz2012} converted through reduction a Sudoku problem into a ``Boolean Satisfiability" problem, also known as SAT. The approach allowed not only the solving, but also the analysis of a Sudoku puzzle difficulty from the polynomial computation time perspective. A similar SAT-based solver was also proposed in \cite{Weber2005}, where the author describes a straightforward translation of a Sudoku grid into a propositional formula. The translation, combined with a general purpose SAT solver was able to solve $9\times 9$ puzzles within milliseconds. In addition, the author suggests that the algorithm can be extended to enumerate all possible solutions for Sudoku grids that are beyond the unique solution grids posed for usual commercial puzzles.

A distinct class of computational solvers is based on stochastic techniques. A solver based on swarm robotics was proposed in \cite{Pacurib2009}. The solver uses an artificial bee colony (ABC) for a guided exploration of the Sudoku grid search space. The algorithm mimics the behaviour of bees when foraging, behaviour which is further used for building partial (local) solutions in problem domain. The purpose of the algorithm is to minimise the number of duplicate digits found on each row and column. The authors compare the ABC algorithm with a Sudoku solver based on a classic genetic algorithm proposed by Mantere \cite{Mantere2007}, and demonstrate that the ABC solver outperforms the GA solver significantly (i.e., on average 6243 processing cycles for ABC, versus 1238749 cycles for GA). In a different study Perez and Marwala \cite{Perez2008arXiv} proposed and compared four stochastic optimisation solvers: a Cultural Genetic Algorithm (CGA), a Repulsive Particle Swarm Optimisation (RPSO) algorithm, a Quantum Simulated Annealing (QSA) algorithm, and a Hybrid method combining a Genetic Algorithm with Simulated Annealing (HGASA). The authors found that the CGA, QSA and HGASA were successful with runtimes of 28, 65 and 1.447 seconds respectively, while the RPSO failed to solve any puzzle. The authors concluded that the very low runtime of HGASA was due to combining the parallel searching of GA with the flexibility of SA. In the same time, they suggested that RPSO was not able to solve the puzzles because the search operations could not be naturally adapted to generating better solutions.

Another class of computational solvers is based on neural networks \cite{Yue2006,Hopfield2008}; however, these solvers are not emphasising on the cognitive plausibility of the neural networks, but rather on their mathematical mechanism. In \cite{Yue2006} the authors propose a Sudoku solver based on the Q'tron energy-driven neural-network model. They map the Sudoku constraints in Qtron's energy function, which is then minimised ensuring the local minimums are avoided through a noise-injection mechanism. The authors show that the algorithm is totally unsuccessful in the absence of noise, while with the noise the success rate is $100\%$ and the runtime is within 1 second. Also they demonstrate that the algorithm can be used not only for solving, but also for generating puzzles with unique solution. In a different approach, Hopfield \cite{Hopfield2008} considers that neural networks do not work well when applied to Sudoku, because they tend to make errors on the way. While \cite{Yue2006} treats this problem by injecting noise in the Q'tron, Hopfield assumes that the search space during a Sudoku game can be mapped into an associative memory which can be used for recognising the inherent errors and reposition the NN representation of the Sudoku grid on the proper search path.

One particular class of computational Sudoku solvers, which is of major interest for our study, is the Constraint Propagation (CP) solvers. Several studies considered that Sudoku puzzle can be treated as a Constraint Satisfaction Problem \cite{Simonis2005,Norvig2014}, and hence, can be solved using constraint programming techniques. Constraint Propagation solvers are purely computational methods, and the studies that proposed them followed the same purpose as the rest of the computational approaches, i.e. to produce proficient Sudoku solvers with minimal computational complexity and no domain knowledge. However, the constraint propagation processes described in both \cite{Simonis2005} and \cite{Norvig2014} are considered to be similar to the steps undertaken by human players when solving Sudoku. In his study \cite{Norvig2014} Norvig emphasises that the major task performed by humans when playing Sudoku is not to fill an empty cell, but to eliminate the multiple candidates for it, as a result of applying and propagating the Sudoku constraints. Yet, Norvig does not mention in which way the propagation of constraints resembles the human thinking. Instead, Simonis \cite{Simonis2005} does, and states that the various Sudoku-related skills used by the human players when trying to eliminate redundant candidates from cells are actually propagation schemes that participate to a constraint propagation process which eventually solves the constraint satisfaction problem. Simonis considers that ``\textit{they [human players] apply complex propagation schemes with names like \textbf{X-Wing} and \textbf{Swordfish} to find solutions of a rather simple looking puzzle called Sudoku. Unfortunately, they are not aware that this is constraint programming}". An even more advanced step towards demonstrating this concept is taken in \cite{Berggren2012} where the authors implement the constraint propagation based on a set of Sudoku skills (e.g. naked candidates, hidden candidates, Nishio-guess). The authors do not relate their algorithm to the constraint propagation formalism, and refer to it as ``rule-based", but they emphasise it ``\textit{consists of testing a puzzle for certain rules that [...] eliminate candidate numbers. This algorithm is similar to the one human solver uses}".

In this study we build on the concepts proposed in the last class of computational Sudoku solvers, and we consider the skill-based approach on constraint propagation problem as central for the skill representation aspect of CRT applied to Sudoku. Thus, in the following section we describe in detail the Sudoku constraints and some of the skills used by human players in solving the game.

\section{The Sudoku game and skills.}

Sudoku is a number puzzle which in its most known form consists of 81 cells contained in a $9\times9$ square grid that is further divided into $9$ boxes of $3\times3$ cells. The aim of the game is to fill all cells in the grid with single digits between 1 and 9, so that a given number appears no more than once in the \textit{unit} it belongs to, where the unit can be a row, a column or a box. These are the Sudoku rules or the constraints. In general the Sudoku problem can be seen as a $n\times n$ grid with $n$ subsequent boxes of $\sqrt{n}\times\sqrt{n}$ cells. The constraints for a grid $G$ can be then expressed in general as follows: 

\begin{enumerate}
	\item \textbf{Cell.} A cell $C_{ij} \in G$ must be filled with exactly one digit $d_{ij}$ with value between $1$ and $n$
	\item \textbf{Row.} All values in a row $i$ must be unique: $d_{ij}\neq d_{ik}$, $\forall i=1,n$ and $\forall j,k=1,n$ with $j\neq k$.
	\item \textbf{Column.} All values in a row $j$ must be unique: $d_{ij}\neq d_{kj}$, $\forall j=1,n$ and $\forall i,k=1,n$ with $i\neq k$.
	\item \textbf{Box.} All values in a box $B_i\in G$ must be unique: $d_{jk}\neq d_{pq}$, $\forall d\in B_i$, with $i=1,n$.
\end{enumerate}

\subsection{Playing a game}

\begin{figure}[h]
    \center
    \subfigure[Sudoku constraints]
    {
        \includegraphics[width=0.45\textwidth]{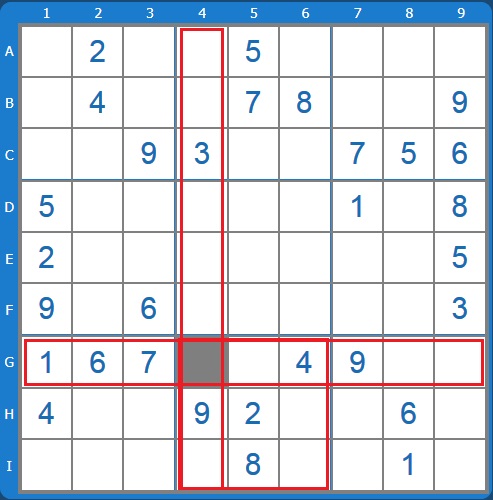}
        \label{fig:Rules}
    }
    ~
    \subfigure[Generate candidates]
    {
        \includegraphics[width=0.45\textwidth]{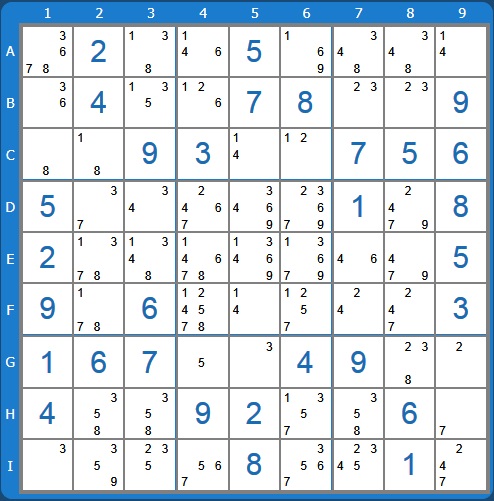}
        \label{fig:Candidates}
    }
    \caption{The game knowledge.}\label{fig:Game}
\end{figure}

In this study we consider the $9\times9$ version of Sudoku. A player applies the Sudoku constraints to empty cells and generates lists of candidates for the visited cells. This process is displayed in Figure~\ref{fig:Game}, where Figure~\ref{fig:Rules} shows the application of rules to cell $C_{G4}$, and Figure~\ref{fig:Candidates} shows the lists of candidates for all empty cells in the grid.

The purpose of the game is to apply Sudoku skills and propagate the Sudoku constraints in order to reduce these lists of candidates to unique candidates \cite{Simonis2005} for all empty cells in the grid, which equals to filling the grid and, thus, solving the puzzle.

\subsubsection{Performing Sudoku skills} \label{PropagationSkill}

In order to reduce the lists of candidates, players use various skills which propagate the domain. In \cite{Aslaksen2014,Davis2008,Chadwick2007} the authors note that players choose the skills based on the perceived context at the current move. The skills considered in this study belong to two categories, the naked and the hidden candidates, which allow the solving of a significant number of Sudoku games. More complex skills \cite{Pitts2010} can be involved for solving very difficult games, however it is outside the scope of this study to investigate an exhaustive list of skills.

\begin{figure}[h]
    \center
    \subfigure[Naked Single]
    {
        \includegraphics[width=0.45\textwidth]{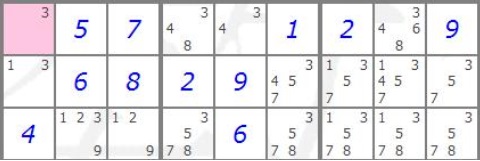}
        \label{fig:NSingle}
    }
    ~
    \subfigure[Naked double]
    {
        \includegraphics[width=0.45\textwidth]{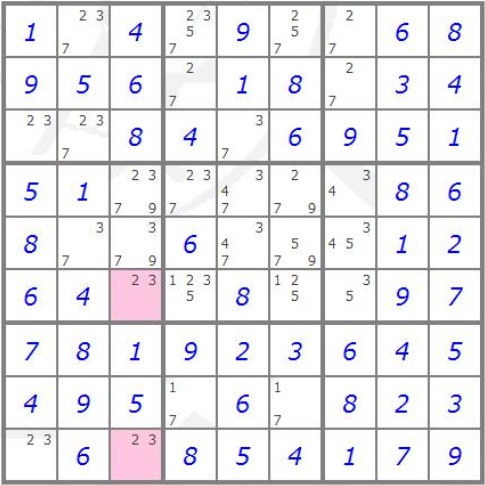}
        \label{fig:NDouble}
    }
    \caption{Naked candidates}\label{fig:Naked}
\end{figure}

The set of naked candidate skills consists of finding and propagating naked singles and doubles (Figure~\ref{fig:Naked}). Recognising and propagating a naked single is the simplest skill, where after the application of Sudoku constraints a cell has only one possible candidate. The value of this unique candidate solves the empty cell, and is propagated by removing the candidate value from the candidate lists of all other cells situated in the units the cell belongs to. A naked single is illustrated in Figure~\ref{fig:NSingle} in pink colour at $C_{A1}$.
For the naked doubles, the lists of candidates are checked for a pair of cells in a Sudoku unit containing only the same two candidates. These candidates can only go in these cells, thus the propagation is done by removing them from the candidate lists of all other unsolved cells in that unit. In Figure~\ref{fig:NDouble} the cells coloured in pink, in column 3 at $C_{F3}$ and $C_{I3}$ show a naked double containing the candidate values (2,3).

\begin{figure}[h]
    \center
    \subfigure[Hidden Single]
    {
        \includegraphics[width=0.45\textwidth]{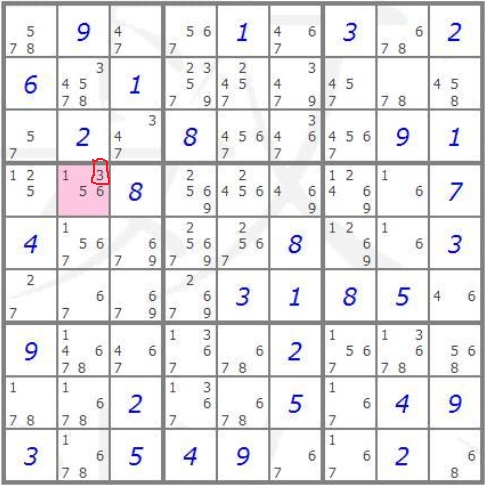}
        \label{fig:HSingle}
    }
    ~
    \subfigure[Hidden double]
    {
        \includegraphics[width=0.45\textwidth]{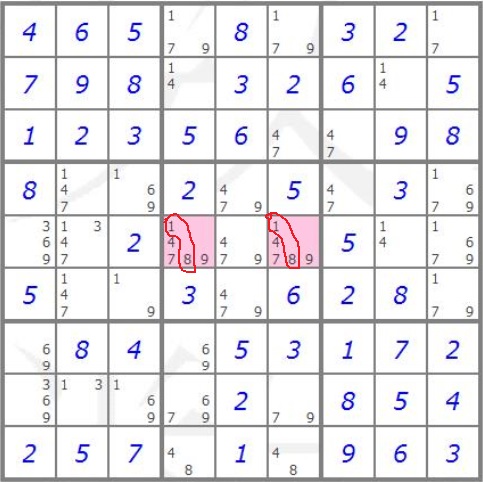}
        \label{fig:HDouble}
    }
    \caption{Hidden candidates}\label{fig:Hidden}
\end{figure}

The set of hidden candidate skills consists of finding and propagating hidden singles and doubles (Figure~\ref{fig:Hidden}). For the hidden single if a candidate value appears in only one cell in a Sudoku unit (row, column or box), the value becomes the unique candidate for that cell, the rest being removed. Thus, the candidate becomes a naked single and further propagates the domain as a naked single. Figure~\ref{fig:HSingle} shows value 3 as a hidden single in cell $C_{D2}$.
For the hidden double, if a given pair of candidates appears in only two empty cells in a unit, then only these candidates must remain in these cells, the other candidates being removed. Thus, the hidden double becomes a naked double and further propagates the domain as a naked double. Figure~\ref{fig:HDouble} shows the hidden pair $(1,8)$ in cells $C_{E4}$ and $C_{E6}$.

\section{Methodology}

In this study we consider that performing skills is subject to pattern recognition, where the player must recognise the pattern of a skill in the lists of candidates in a unit, in order to be able to apply that skill. We model the acquisition of skills through supervised training of feed-forward neural networks with back-propagation mechanism, one network for each skill.
We treat two possible situations in skill acquisition. First, we train the ability to recognise the existence of a skill pattern in a Sudoku unit (cell, column or box) and we call this case ``skill detection". Second, we train the ability to recognise not only the existence of a skill pattern, but also the cells in the unit which the skills is applicable to. We call this case ``skill localisation". In the two cases the resultant neural networks have similar number of neurons in the input and hidden layer, and similar training sets $x$ for learning the skills, but they have different number of neurons in the output layer and, consequently, different target sets $t$.

\begin{figure}[h]
\begin{center}
\includegraphics[width=0.8\linewidth]{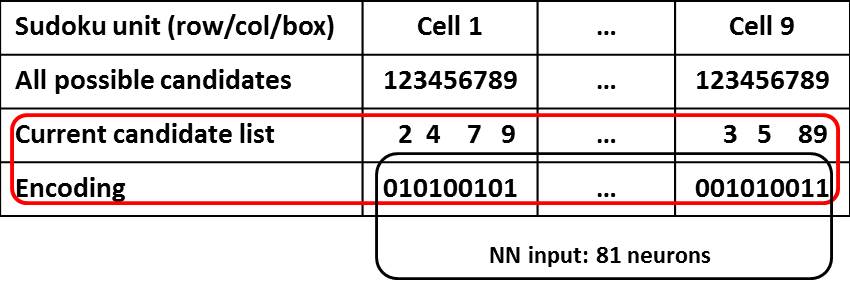}
\caption{Neural network input}
\label{fig:Encoding}
\end{center}
\end{figure}

Figure~\ref{fig:Encoding} shows the encoding of candidate list information into the input layer of neural networks. In a Sudoku unit, each of the nine cells can have a maximum of nine potential candidates, i.e. the digits from $1$ to $9$. However, at a certain step in the game the current candidate lists usually contain less than nine digits; the lists can be depicted as in the third row of the table. We encode the decimal values of the candidates into binary values as presented in the third row. The total length of the binary encoded lists of candidates is 81, thus, we use neural networks with 81 neurons in the input layer.

\begin{figure}[h]
    \center
    \subfigure[Skill detection]
    {
        \includegraphics[width=0.45\textwidth]{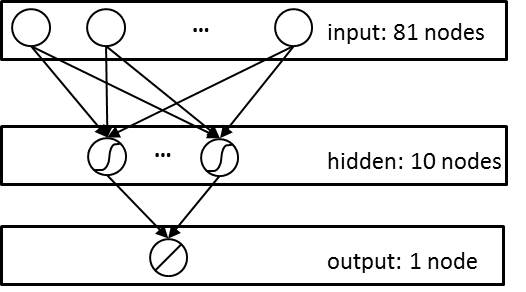}
        \label{fig:output1}
    }
    ~
    \subfigure[Skill localisation]
    {
        \includegraphics[width=0.45\textwidth]{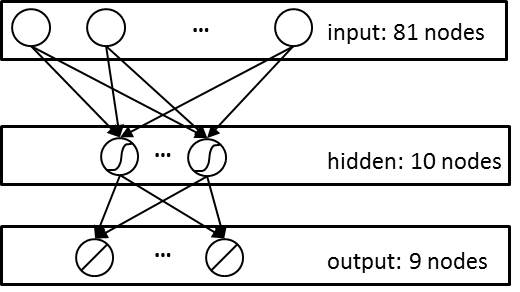}
        \label{fig:output9}
    }
    \caption{Neural networks for skill pattern recognition}\label{fig:NeuralNet}
\end{figure}

\subsection{Skill detection}

For detecting the states we adopt the network structure presented in Figure~\ref{fig:output1}, with one node in the output layer which shows if the pattern of a skill $S_i$ is present in a Sudoku unit.
For each skill we use artificially generated training and target sets, as presented in Algorithm~\ref{Alg:TrainingSet1}. For the skills treated in this study a training sample $x_i$ is a binary vector with 81 elements corresponding to the 81 nodes in the input layer. 

\subsubsection{Single candidates}

The training dataset is a binary matrix consisting of 162 samples ($X_{ij}, i=1:162, j=1:81$). 81 samples correspond to all possible appearances of a naked or hidden single in a unit (i.e. there can be 81 naked single situations in a Sudoku column, row or box), for which the values of the target set $t(1:1296)=1$. The other 81 samples do not contain the single candidate pattern, hence the values of the target set are $t(82:162)=-1$.

\subsubsection{Double candidates}

The training dataset for double candidate skills (naked and hidden doubles) is a binary matrix consisting of 2592 samples ($X_{ij}, i=1:2592, j=1:81$). 1296 samples correspond to all possible appearances of a skill in a unit (i.e. there can be 1296 naked double situations in a Sudoku column, row or box), for which the values of the target set $t(1:1296)=1$. The other 1296 samples do not contain the skill pattern, hence the values of the target set are $t(1297:2592)=-1$.

\begin{algorithm}[H]
\caption{Skill detection: training and target sets for a skill $S_i \in SkillSet$}
\label{Alg:TrainingSet1}
\scriptsize
\begin{algorithmic}[1]
    \STATE \COMMENT {Input: Skill $S_i$}
    \FOR {i = 1 \TO No. of $S_i$ patterns in a Sudoku unit}
    	\FORALL {cells in the Sudoku unit}
			\STATE {trainingSet: $x(i,allcells)= S_i pattern$}
    	\ENDFOR
    	\STATE {targetSet: $t(i)=1$}
    \ENDFOR
    \FOR {j = 1 \TO No. of $S_i$ patterns in a Sudoku unit}
    	\FORALL {cells in the Sudoku unit}
			\STATE {trainingSet: $x(i+j,allcells)= random pattern$}
    	\ENDFOR
    	\STATE {targetSet: $t(i+j)=-1$}
    \ENDFOR
\end{algorithmic}
\end{algorithm}

\subsection{Skill localisation}

For locating the patterns of skills we adopt the network structure presented in Figure~\ref{fig:output9}, with nine nodes in the output layer. The nine nodes correspond to the nine cells in a Sudoku unit. Depending on which skill is subject to recognition, the cells in which the skill pattern exists will fire.
The training dataset for this case is generated in a similar manner to the previous case. The generation is presented in Algorithm~\ref{Alg:TrainingSet9}, where the training matrix $X$ is similar to that from the skill detection case ($X_{ij}, i=1:2592, j=1:81$). The target set is a matrix $T(i,k)$ with ($i=1:TrainSetSize,k=1:9$), defined as in Equation~\ref{eq:TargetMatrix}.

\begin{equation}
\label{eq:TargetMatrix}
  T(i,k) =
   \begin{dcases*}
     1 & if $X(i,1:81)$ contains the skill pattern \\
    -1 & otherwise
   \end{dcases*}
\end{equation}

\begin{algorithm}[H]
\caption{Skill localisation: training and target sets for a skill $S_i \in SkillSet$}
\label{Alg:TrainingSet9}
\scriptsize
\begin{algorithmic}[1]
    \STATE \COMMENT {Input: Skill $S_i$}
    \FOR {i = 1 \TO No. of $S_i$ patterns in a Sudoku unit}
    	\FORALL {cells in the Sudoku unit}
			\STATE {trainingSet: $x(i,allcells)= S_i pattern$}
			\STATE {targetSet: $t(i,allOutputNodes)=1$}
    	\ENDFOR
    \ENDFOR
    \FOR {j = 1 \TO No. of $S_i$ patterns in a Sudoku unit}
    	\FORALL {cells in the Sudoku unit}
			\STATE {trainingSet: $x(i+j,allcells)= random pattern$}
			\STATE {targetSet: $t(i,allOutputNodes)=-1$}
    	\ENDFOR
    \ENDFOR
\end{algorithmic}
\end{algorithm}

\subsection{Network and training settings}

We use the standard $tanh$ for activation function of nodes in the networks and the mean square root error function (MSE) for the subsequent gradient minimisation. The artificially generated training sets are split in ratios of 0.7, 0.15 and 0.15 for training, internal cross validation and generalisation testing, respectively.

\subsection{Skill aggregation - the solver}

The constraint propagation side of the Sudoku solving is hard-coded. However, the recognition of the patterns for each of the four skills considered in the study is implemented using the neural networks, and hence the ability to recognise either the existence of a skill pattern (detection) or its location (localisation) depends on the ability of the neural networks to produce the desired output.  This implementation, with the hard-coded solving sequence, and the NN representation of the skills is error free from the Sudoku solving point of view, since it avoids situations when multiple states coexist in one board, i.e. a single candidate and double candidate simultaneously. Since the networks we propose are only meant to demonstrate the individual skills, they cannot treat cases where a combination of skills is present, or the player must choose from multiple skills. Since this aspect was outside of the scope of this study, we adopted a predetermined solving sequence implemented in the hard-coded constraint propagation module.

\section{Results and discussion}

\begin{figure}[h]
    \center
    \subfigure[Naked single]
    {
        \includegraphics[width=0.45\textwidth,height=0.2\textheight]{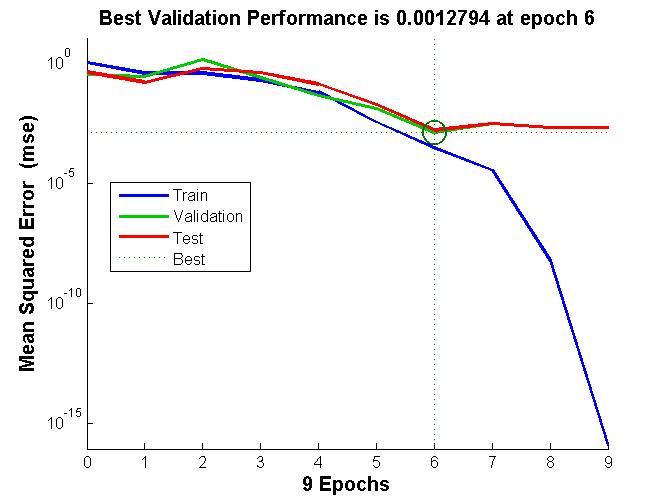}
        \label{fig:NSingleTrainState}
    }
    ~
    \subfigure[Naked Double]
    {
        \includegraphics[width=0.45\textwidth,height=0.2\textheight]{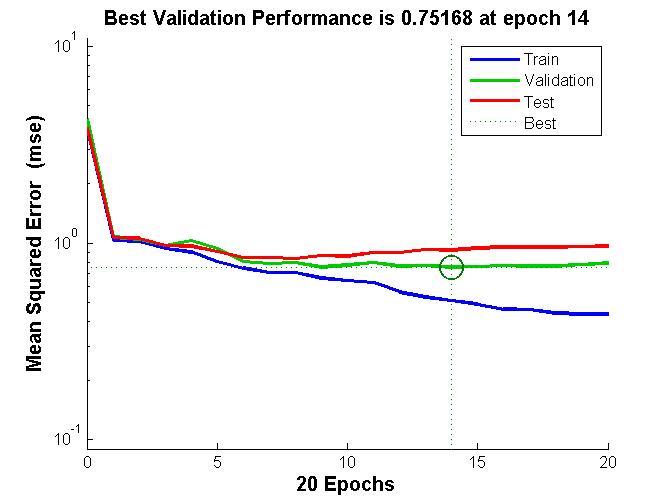}
        \label{fig:NDoubleTrainState}
    }
    \subfigure[Hidden Single]
    {
        \includegraphics[width=0.45\textwidth,height=0.2\textheight]{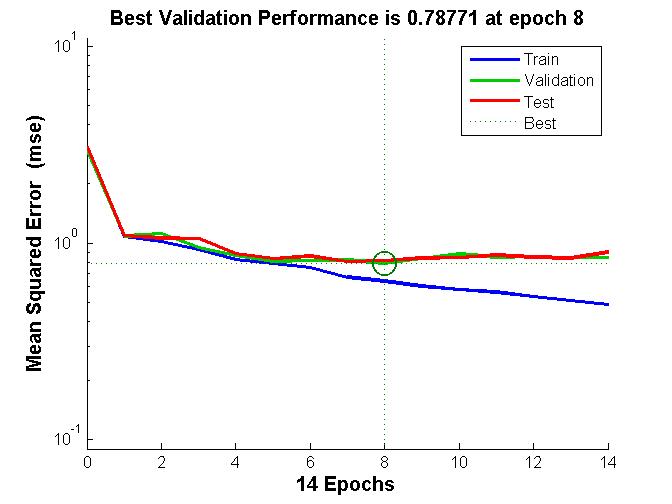}
        \label{fig:HSingleTrainState}
    }
    ~
    \subfigure[Hidden Double]
    {
        \includegraphics[width=0.45\textwidth,height=0.2\textheight]{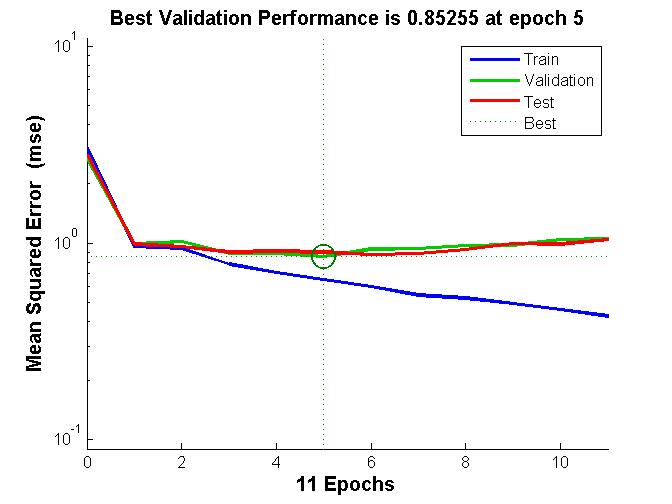}
        \label{fig:HDoubleTrainState}
    }
    \caption{NN training for skill detection.}\label{fig:SkillDetect}
\end{figure}

\begin{figure}[h]
\centering
    \includegraphics[width=0.75\linewidth]{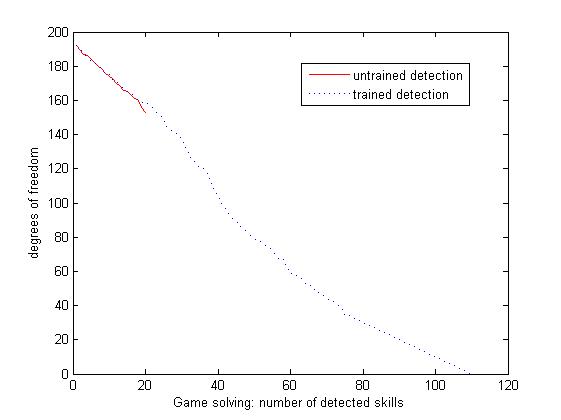}
    \caption{Skill detection. Sudoku solving with trained and untrained skills.}
    \label{fig:SkillDetectComp}
\end{figure}

Figure~\ref{fig:SkillDetect} shows the results of the training process in the skill detection recognition case. The training of each of the four skills is considered finished when the best validation performance is reached.
Table~\ref{table:SkillDetect} and Figure~\ref{fig:SkillDetectComp} present the game solving results for both trained and untrained skills situations, where the untrained skills are the skills acquired after one epoch in the neural networks. Results demonstrate how the proficiency of the skill-based solver improves with the acquisition of skills. In the table the difference between the number of detected skill patterns is shown for the two cases, while in the figure the result of game solving is shown in terms of the degrees of freedom. We demonstrate that the NN-based skill detection training is able to solve the proposed Sudoku game, provided that the rest of the solving mechanism is hard-coded in the solver.

\begin{table}
	\center{
    	    \begin{tabular}{|l|c|c|c|c|c|}\hline
       	         & Number of & Number of & Number of & Number of & Game result \\
       	         & naked singles & naked doubles & hidden singles & hidden doubles & degree of freedom \\ \hline
    	        untrained & 2 & 0 & 17 & 1 & 153 \\ \hline
        	    trained & 54 & 5 & 50 & 1 & 0 \\ \hline
        	\end{tabular}
	        }
	\caption{Skill detection: Sudoku solving competency.}
	\label{table:SkillDetect}
\end{table}

\begin{figure}[b]
    \center
    \subfigure[Naked single]
    {
        \includegraphics[width=0.45\textwidth,height=0.2\textheight]{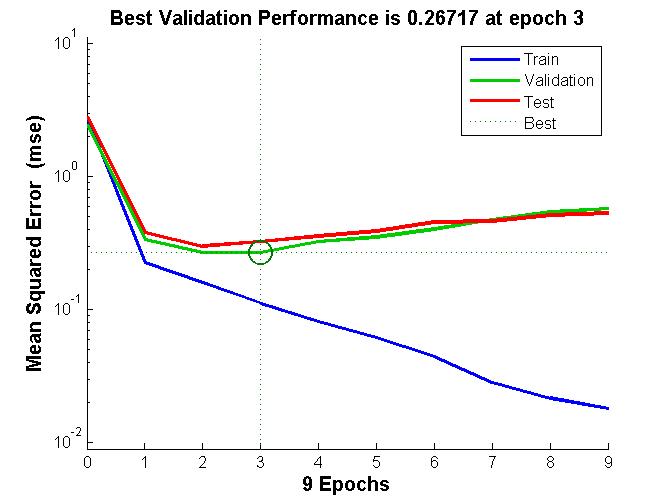}
        \label{fig:NSingleTrainSkill}
    }
    ~
    \subfigure[Naked Double]
    {
        \includegraphics[width=0.45\textwidth,height=0.2\textheight]{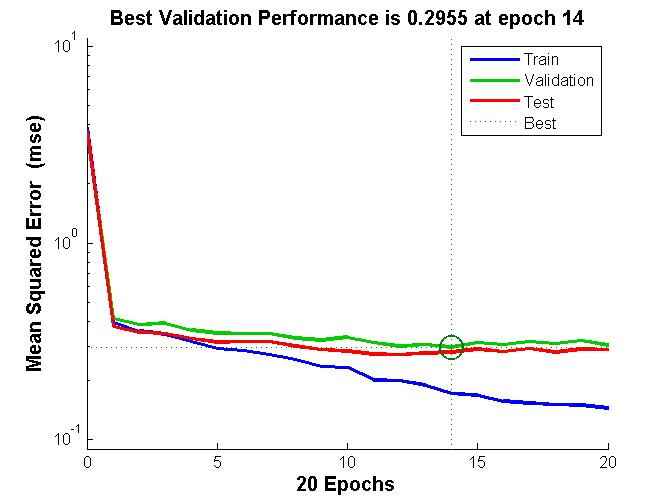}
        \label{fig:NDoubleTrainSkill}
    }
    \subfigure[Hidden Single]
    {
        \includegraphics[width=0.45\textwidth,height=0.2\textheight]{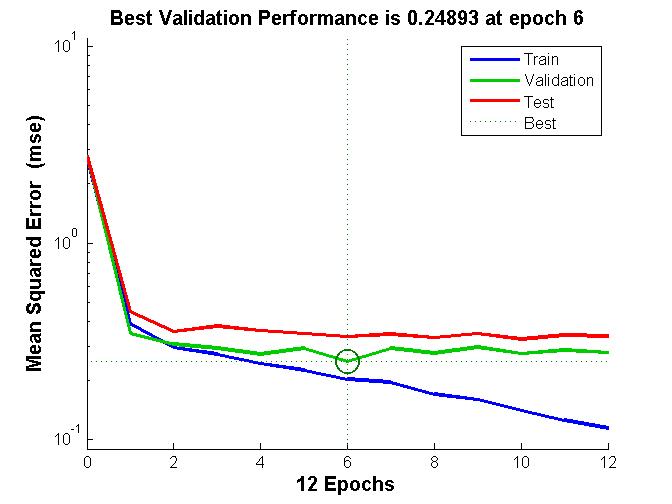}
        \label{fig:HSingleTRainSkill}
    }
    ~
    \subfigure[Hidden Double]
    {
        \includegraphics[width=0.45\textwidth,height=0.2\textheight]{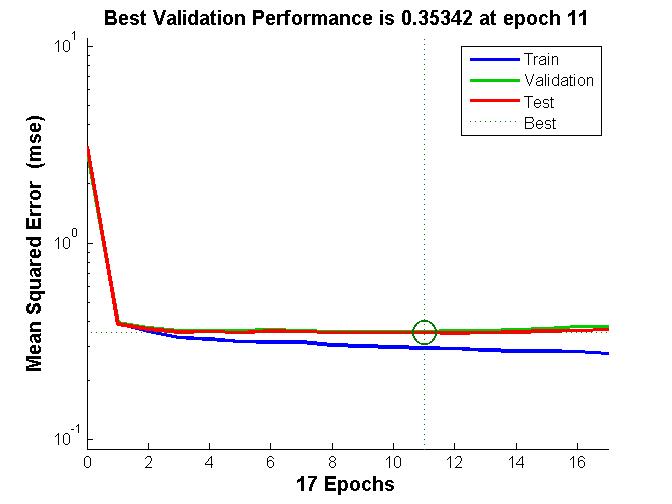}
        \label{fig:HDoubleTrainSkill}
    }
    \caption{NN training for skill localisation.}\label{fig:SkillLoc}
\end{figure}

\begin{figure}[b]
\centering
    \includegraphics[width=0.75\linewidth]{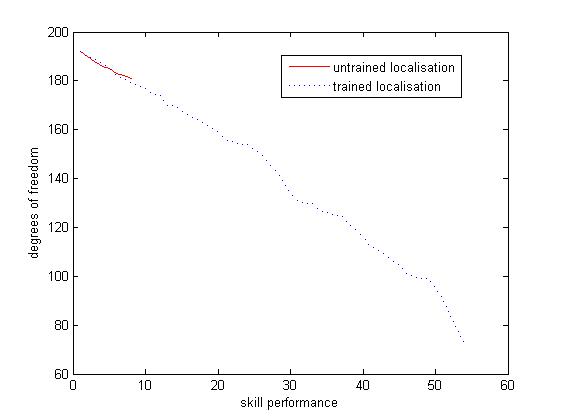}
    \caption{Skill localisation. Sudoku solving with trained and untrained skills.}
    \label{fig:SkillLocComp}
\end{figure}

For the skill localisation case, results of the training process are shown in Figure~\ref{fig:SkillLoc}. Similar to the skill detection case, the training of each of the four skills is considered finished when the best validation performance is reached. Table~\ref{table:SkillLoc} and Figure~\ref{fig:SkillLocComp} present the results of game solving for trained and untrained skills, where the untrained skills are the skills acquired after one epoch in the neural networks. Results demonstrate again that the proficiency of the skill-based solver improves with the acquisition of skills. The proficiency in this case is lower, a result which is expected given that the solver must recognise not only the existence of a skill in a unit, but also the skill pattern. Results show an improvement in the number of recognised skills, which subsequently leads to less degrees of freedom, but the solver still does not reach the end of the proposed Sudoku game. However, since the proficiency is not the purpose of this study, we emphasise on the improvement resulted from skill acquisition using NN training.

\begin{table}
	\center{
    	    \begin{tabular}{|l|c|c|c|c|c|}\hline
       	         & Number of & Number of & Number of & Number of & Game result \\
       	         & naked singles & naked doubles & hidden singles & hidden doubles & degree of freedom \\ \hline
    	        untrained & 0 & 1 & 6 & 1 & 181 \\ \hline
        	    trained & 31 & 3 & 19 & 1 & 73 \\ \hline
        	\end{tabular}
	        }
	\caption{Skill localisation: Sudoku solving competency.}
	\label{table:SkillLoc}
\end{table}

\section{Conclusions}

In this paper we focused on the skill assessment aspect of the CRT process, for which the representation of skills is a central and essential issue. We investigated this using the Soduku puzzle by introducing a plausible representation of Sudoku skills, and by modelling the process of acquiring these skills. We used feed-forward neural networks with back-propagation mechanism for training the skills and we tested the resultant skills in a cognitively plausible skill-based computational Sudoku solver.

The results of Sudoku game solving demonstrated the plausibility of using skills in computational Sudoku solvers, and also demonstrated the concept of skill acquisition in relation to the proficiency of this solver. We found that a skill-based computational Sudoku solver can achieve certain levels of proficiency by learning the Sudoku skills using neural networks. Results are encouraging for developing more complex skill and skill acquisition models usable in more general frameworks related to skill assessment stage of the Computational Red Teaming process.

\section*{Acknowledgement}

This project is supported by the Australian Research Council Discovery Grant DP140102590, entitled ``Challenging systems to discover vulnerabilities using computational red teaming". \\
This is a pre-print of an article published in Proceedings in Adaptation, Learning and Optimization, vol 5, Springer. The final authenticated version is available online at: https://doi.org/10.1007/978-3-319-27000-5\_26

\clearpage
\footnotesize
\bibliographystyle{splncs03}

\end{document}